\documentclass[runningheads]{llncs}
\usepackage[T1]{fontenc}
\usepackage{graphicx}
\usepackage{caption}
\usepackage{cite}
\usepackage{amsmath}
\usepackage{amssymb}

\usepackage[]{algorithm}
\usepackage{algpseudocode}
\algnewcommand\And{\textbf{and}}
\usepackage{amsfonts}
\usepackage{rotating}
\usepackage{xcolor}
\usepackage{multirow}
\usepackage{color}
\usepackage{array}
\usepackage{booktabs}
\usepackage{tabularx}
\usepackage{multirow}
\usepackage{placeins}
\usepackage{longtable}
\usepackage{supertabular,enumitem}
\usepackage{inconsolata}
\usepackage{hyperref}
\usepackage{academicons}
\definecolor{orcidlogocol}{HTML}{A6CE39}
\usepackage{xspace}
\usepackage{todonotes}
\usepackage{booktabs}
\usepackage{amsmath}
\usetikzlibrary{arrows.meta}
\usepackage{eurosym}
\usepackage{diagbox}
\usepackage{subcaption}

\definecolor{inteins}{RGB}{128,179,255}

\newcommand{\calL}{\mathcal{L}}

\newcommand{\calR}{\mathbb{R}}
\newcommand{\calE}{\mathbb{E}}
\newcommand{\MNIST}{\texttt{MNIST}\xspace}
\newcommand{\RING}{\texttt{RING}\xspace}
\newcommand{\BLOB}{\texttt{BLOB}\xspace}

\newcommand{\methodLong}{Co-evolutionary Elitist SSL-GAN}
\newcommand{\method}{CE-SSLGAN}
\begin{document}
\title{Generate more than one child in your co-evolutionary semi-supervised learning GAN}
\titlerunning{Generate more than one child in your co-evolutionary SSL-GAN}
%
\author{Francisco Sedeño\orcidID{0009-0000-7797-3562} \and
Jamal Toutouh\orcidID{0000-0003-1152-0346} \and
Francisco Chicano\orcidID{0000-0003-1259-2990}}
\authorrunning{F. Sedeño et al.}
%
\institute{ITIS Software, University of Malaga, Spain \\
\email{\{fsedenoguerrero,jamal,chicano\}@uma.es}}
\maketitle              
\begin{abstract}
Generative Adversarial Networks (GANs) are very useful methods to address semi-supervised learning (SSL) datasets, thanks to their ability to generate samples similar to real data. This approach, called SSL-GAN has attracted many researchers in the last decade. Evolutionary algorithms have been used to guide the evolution and training of SSL-GANs with great success. In particular, several co-evolutionary approaches have been applied where the two networks of a GAN (the generator and the discriminator) are evolved in separate populations. The co-evolutionary approaches published to date assume some spatial structure of the populations, based on the ideas of cellular evolutionary algorithms. They also create one single individual per generation and follow a generational replacement strategy in the evolution. In this paper, we re-consider those algorithmic design decisions and propose a new co-evolutionary approach, called \emph{\methodLong} (\method), with panmictic population, elitist replacement, and more than one individual in the offspring. We evaluate the performance of our proposed method using three standard benchmark datasets. The results show that creating more than one offspring per population and using elitism improves the results in comparison with a classical SSL-GAN. 
\end{abstract}
\section{Introduction}
\label{sec:introduction}


A \emph{generative adversarial network} (GAN)~\cite{DBLP:conf/nips/GoodfellowPMXWOCB14} is composed of two artificial neural networks: a \emph{generator network} and a \emph{discriminator network}. The generator creates data samples trying to follow the probability distribution of the real data samples for a particular application, while the discriminator tries to distinguish between real or generated (fake) data samples. The loss functions used to train the networks are defined in such a way that: 1) the generator is rewarded when it generates data samples the discriminator classifies as real and 2) the discriminator is rewarded when it correctly distinguishes between real and fake data samples.
GANs are able to produce a discriminator for a particular application using less samples than other neural networks due to the generative nature of the network. The generator is able to produce realistic data samples and, for this reason, GANs have been used in medical image generation~\cite{DBLP:journals/asc/WangWY21}, 3D object generation~\cite{DBLP:journals/jise/ByrdDCCK22}, and image-to-image translation~\cite{DBLP:journals/asc/QinWX22} among others.

GAN training is not an easy task. Vanishing gradient, mode collapse or discriminator collapse are some of the problems that can be found during the training. In order to mitigate these problems, researchers have successfully proposed the use of evolutionary algorithms (EA) with a co-evolution approach~\cite{DBLP:journals/asc/ToutouhNHO23}. One prominent example is Lipizzaner~\cite{DBLP:journals/corr/abs-1811-12843}, which combines co-evolution with a spatial distribution of the individuals in the population borrowed from cellular EAs~\cite{DorronsoroAlba2008}.

The ability of GANs to learn from less labeled data makes them very appropriate for semi-supervised learning (SSL), where the datasets contain a mix of labeled and unlabeled data samples. The use of adversarial networks in SSL, sometimes denoted with the acronym SSL-GAN, is very old~\cite{DBLP:conf/ACISicis/TachibanaMU16} and there are many different proposals in the scientific literature. The interested reader is referred to the recent surveys by Sajun and Zualkernan~\cite{app12031718} and Ma et al.~\cite{Ma2024}.
The application of co-evolutionary approaches to SSL-GANs is recent, with only two works to date~\cite{DBLP:conf/gecco/ToutouhNHO23,DBLP:journals/asc/ToutouhNHO23}, up to authors knowledge. The previous work on co-evolutionary SSL-GANs propose the use of a particular spatial distribution of the individuals, one offspring per generation, and a generational replacement strategy for the population.

In this work, we present a novel co-evolutionary approach for SSL-GANs and evaluate different parameters of the approach. 
The novel features of the proposal are the use of panmictic population, elitist replacement, and more than one individual in the offspring.
The final goal of this study is to create a guide that could be useful in future studies to set the parameters of a co-evolutionary SSL-GAN.

The remainder of the paper is organized as follows. Section~\ref{sec:background} presents the definitions and background information required to understand the rest of the paper. Our novel SSL-GAN approach is presented in Section~\ref{sec:method}. We propose our research questions and describe the experimental methodology in Section~\ref{sec:experimental-setup}. Finally, Section~\ref{sec:results} presents the results obtained in the experiments and the paper concludes in Section~\ref{sec:conclusions}.


\section{Background}
\label{sec:background}

Let us assume that we have a partially labeled real samples dataset containing samples classified in $K$ classes.
We assume that the samples are $d$-dimensional real vectors $x \in \calR^{d}$. For each labeled real sample we have its class represented using one-hot encoding $y \in \{0,1\}^K$. We will denote the labeled real samples with $p_{real, la} \subseteq \calR^d \times \{0,1\}^K$, and the unlabeled real samples with $p_{real, un} \subseteq \calR^d$.

\begin{figure}[!ht]
    \centering
    \includegraphics[width=0.65\linewidth]{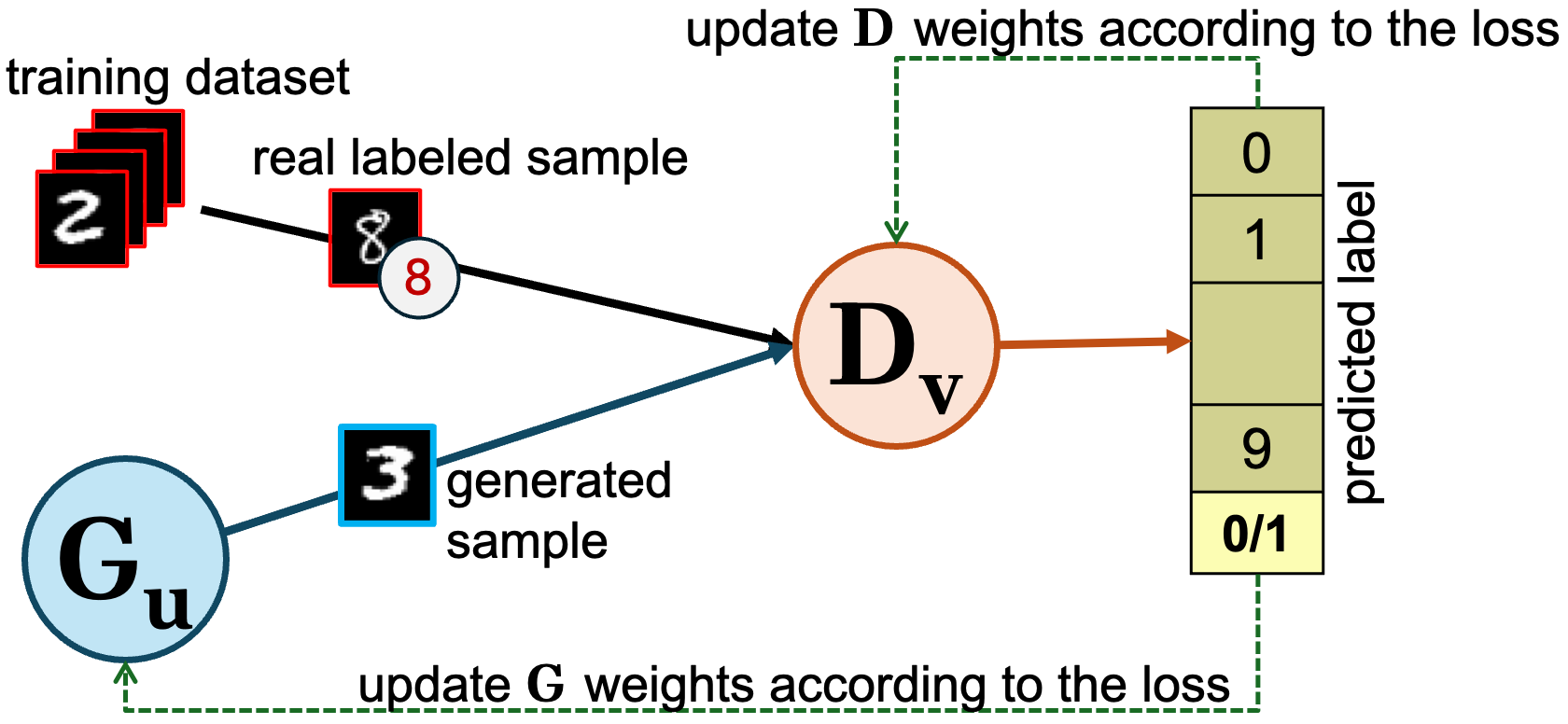}
    \caption{
    Example of SSL-GAN for MNIST digits generation}
    \label{fig:ssl-gan-representation}
\end{figure}

An SSL-GAN (Fig.~\ref{fig:ssl-gan-representation}) is composed of two artificial neural networks (ANN): a generator, $G_u$, which generates a data sample from a random input vector; and a discriminator, $D_v$, which classifies each sample as real or fake and provides one of the $K$ classes for real data.
The $u$ in $G_u$ and the $v$ in $D_v$ denote the set of parameters of the generator and discriminator, respectively. Formally, $G_u$ is a function $G_u: \calR^{\ell} \rightarrow \calR^{d}$ that maps a vector in $\calR^{\ell}$, called the latent space, to a generated sample in $\calR^{d}$.
The discriminator $D_v$ is a function $D_v: \calR^{d} \rightarrow [0,1]^{K+1}$ that maps a sample to a $K+1$-dimensional vector of real values in $[0,1]$. The first $K$ values of that vector are probabilities for the $K$ classes in the real data and we will denote them as $D_v^{class}(x) \in [0,1]^{K}$. The $(K+1)$-th value in the vector is the probability for the sample to be real and we will denote it as $D_v^{real}(x) \in [0,1]$, where $D_v^{real}(x)=1$ means that sample $x$ is classified as real with certainty and $D_v^{real}(x)=0$ means it is classified as fake with certainty.

The goal of the generator is to generate samples similar to the real samples. Thus, its loss function should take lower values when $D_v^{real}(G_u(z))$ is higher. The loss function for the generator can be written as:
\begin{equation}
\label{eq:sgan_generator}
\calL_{G} = \calE_{z\sim \mathcal{N}^{\ell}(\mathbf{0}, \mathbf{I})}[\phi(D_{v}^{real}(G_u(z)))], 
\end{equation}

where $\phi$ is a monotone decreasing function and $\mathcal{N}^{\ell}(\mathbf{0}, \mathbf{I})$ is a $\ell$-dimensional multivariate normal distribution with zero mean and identity covariant matrix (all components of the vector are independent and follow a standard normal distribution). In our case, we use cross-entropy for the loss function, and $\phi(y) = - \log y$.

The discriminator has two goals: 1) classifying each sample as real or fake and 2) for real data, classifying the sample in one of the $K$ classes. The loss function of the discriminator is
\begin{equation}
\label{eq:sgan_discriminator}
\calL_{D} = \calL_{D,u} + \calL_{D,s},
\end{equation}
where $\calL_{D,u}$ is the \emph{unsupervised loss}, based on the fake samples and unlabeled real data, and $\calL_{D,s}$ is the \emph{supervised loss}, based on the real labeled data. These losses are
\begin{align}
\label{eq:sgan_d_unsupervised}
\calL_{D,u} &= \calE_{z\sim \mathcal{N}^{\ell}(\mathbf{0}, \mathbf{I})}\left[\phi(1-D_{v}^{real}(G_u(z)))\right]
+\calE_{x \sim p_{real, un}}\left[ \phi (D_{v}^{real}(x)) \right],\\
\label{eq:sgan_d_supervised}
\calL_{D,s} &= \calE_{(x, y) \sim p_{real, la}}\left[ \sum_{i=1}^{K} y_i \phi (D_{v,i}^{class}(x)) \right],
\end{align}

where we used $D_{v,i}^{class}(x)$ to denote the $i$-th component of the output vector $D_v^{class}(x)$ in the discriminator. 

\subsection{SSL-GAN training}
\label{sec:sslgan-method}

SSL-GAN extends the traditional unsupervised GAN by introducing an additional objective for the discriminator: classifying the real labeled images. Algorithm~\ref{alg:sslgan-training} summarizes the SSL-GAN training process.
The training proceeds over $T$ epochs, iterating through batches of labeled, $B_L$, and unlabeled, $B_U$, real samples to compute $\mathcal{L}_{D,s}$, $\mathcal{L}_{D,u}$, $\mathcal{L}_{D}$, and $\mathcal{L}_{G}$.
The two loss functions, $\mathcal{L}_{D}$ and $\mathcal{L}_{G}$, are used to update the network weights of the discriminator (Lines~\ref{alg:sa-d-training-d_ini} to~\ref{alg:sa-d-updating-d}) and the generator (Lines~\ref{alg:sa-g-training-g_ini} to~\ref{alg:sa-g-training-g_fin}), respectively, by stochastic gradient descent.

\begin{algorithm}[h!]
  \small
  \captionsetup{font=small}
  \caption{SSL-GAN training\newline
    \textbf{Input:}
    $T$: Epochs, 
    ~$p_{real,la}$: Dataset with labeled data,
    ~$p_{real,un}$: Dataset with unlabeled data,
    ~$B_s$: Batch size, 
    \hbox{$\theta_{g}$, $\theta_{d}$: Initial generator and discriminator parameters}
    \newline
    \textbf{Return:}
    $g,d$: Trained generator and discriminator
  }\label{alg:sslgan-training}
  \begin{algorithmic}[1]
    \State $g \gets$ initializeNetworks($\theta_{g}$)\Comment{Initialize the generator}
    \State $d \gets$ initializeNetworks($\theta_{d}$)\Comment{Initialize the discriminator}
    \For{$t=1$ \textbf{to} $T$}  \Comment{Loop over the $T$ training epochs}\label{alg:loop-training}
      \For{$B_L \subseteq p_{real,la}, B_U \subseteq p_{real,un}$} \Comment{Loop over the batches in the dataset}\label{alg:loop-batches}
        \State $\mathbf{z} \gets$ getRandomNoise($B_s$)\Comment{Get samples from latent space, $\mathbf{z} \sim \mathcal{N}^{\ell}(\mathbf{0}, \mathbf{I})$}\label{alg:sa-d-training-d_ini}
        \State $X_f \gets$ $g$($\mathbf{z}$)\Comment{Generate fake samples}\label{alg:sa-d-creating-g}
        \State $\mathcal{L}_{D,s} \gets$ getSupervisedLoss($d, B_L$)\Comment{Compute supervised loss}\label{alg:d-loss-d-sup}
        \State $\mathcal{L}_{D,u} \gets$ getUnsupervisedLoss($d, g, B_U, X_f$)\Comment{Compute unsupervised loss}\label{alg:d-loss-d-unsup}

        \State $\mathcal{L}_{D} \gets \mathcal{L}_{D,s}+ \mathcal{L}_{D,u}$\Comment{Compute total discriminator loss}\label{alg:sa-d-loss-total}
        \State $d \gets$ gradientDescent($d, \mathcal{L}_{D}, \theta_{d}$)\Comment{Update discriminator parameters}\label{alg:sa-d-updating-d}
        
        \State $\mathbf{z} \gets$ getRandomNoise($B_s$)\Comment{Get samples from latent space, $\mathbf{z} \sim \mathcal{N}^{\ell}(\mathbf{0}, \mathbf{I})$}\label{alg:sa-g-training-g_ini}
        \State $X_f \gets$ $g$($\mathbf{z}$)\Comment{Generate fake samples}\label{alg:sa-g-creating-g}
        \State $\mathcal{L}_{G} \gets$ generatorLoss($d, g, X_f$)\Comment{Compute generator loss}\label{alg:sa-g-loss-d}
        
        \State $g \gets$ gradientDescent($g, \mathcal{L}_{G}, \theta_{g}$)\Comment{Update generator parameters}\label{alg:sa-g-training-g_fin}
      \EndFor
    \EndFor
    \State \Return $g, d$ \Comment{Return trained $g$ and $d$}
  \end{algorithmic}
\end{algorithm}

\section{\methodLong}
\label{sec:method}


This section presents \method, a competitive co-evolutionary method to train SSL-GANs with limited resources, using small populations. \method\ combines a $(\mu + \lambda)$ competitive co-evolutionary algorithm with the SSL-GAN training approach (see Section~\ref{sec:background}).
\method\ operates over two populations of ANNs of the same size $\mu$: a population of generators $\mathbf{g}$, $\{g_1, \ldots, g_\mu\}$, and a population of discriminators $\mathbf{d}$, $\{d_1, \ldots, d_\mu\}$. In each generation, a subset of $\lambda$ individuals (ANNs) of $\mathbf{g}$ and $\mathbf{d}$ are selected to evolve.
The variation operator of this co-evolutionary algorithm is the mutation of the parameters (weights) of the ANNs, which is performed by applying stochastic gradient descent during a training process. This training is applied to pairs of individuals $g, d$ of both populations by applying a number $n_t$ of training epochs of Algorithm~\ref{alg:sslgan-training}.
Thus, given a computational budget $T_B$ of maximum training epochs, the number of the generations, $\iota$, of the main co-evolutionary loop is given by the expression
\begin{equation}
\iota = \left\lfloor\frac{T_B}{n_t \lambda} \right\rfloor .
    \label{eq:generations}
\end{equation}

Algorithm~\ref{alg:our-method} describes the main steps of \method. 
The algorithm begins by initializing each individual, i.e., ANN, of the generator and discriminator  populations using parameters $\theta_{g}$ and $\theta_{d}$, respectively (Lines~\ref{alg:initialization-g} and~\ref{alg:initialization-d}). The number of generations $\iota$ is computed. Then, all the individuals of both populations are evaluated. The fitness values $\mathcal{L}_{g,d}$ are computed by aggregating the loss of each individual against all the individuals of the adversary population (Line~\ref{alg:fit-eval-1}). A number $n_e$ of batches is used to calculate the loss. The selected value for $n_e$ can affect the individual evaluation, because a small $n_e$ can produce imprecise loss estimations, whereas a large $n_e$ could incur significant computational costs. 

The main \method\ loop runs over $\iota$ generations. 
In each generation, tournament selection with size $\tau$ is applied to create an offspring of generators ($\mathbf{g}^*$) and discriminators ($\mathbf{d}^*$) with size $\lambda$ (Lines~\ref{alg:select-g} and \ref{alg:select-d}). 
The individuals of the offspring are coupled in $\lambda$ SSL-GAN to perform the training for $n_t$ epochs (Line~\ref{alg:method-ssl-training}). 
After training, $\mathbf{g}^*$ and $\mathbf{d}^*$ are added to their populations (Lines~\ref{alg:add-offspring-g} and~\ref{alg:add-offspring-d}, respectively). The updated populations are then evaluated to update the fitness values, $\mathcal{L}_{g,d}$ (Line~\ref{alg:fit-eval-2}).
The best $\mu$ individuals in each population is selected to be the population for the next generation (Lines~\ref{alg:update-pops-g} and~\ref{alg:update-pops-d}).
Once all generations are completed, \method\ returns the  best generator and discriminator according to $\mathcal{L}_{g,d}$ (Lines~\ref{alg:select-best-g} to~\ref{alg:return}).

\begin{algorithm}[!ht]
	\footnotesize
        \captionsetup{font=small}
	\caption{\method\ training\newline
		\textbf{Input:}
            $T_B$: Computational budget (training epochs), 
            $p_{real,la}$: Dataset with labeled data,
            $p_{real,un}$: Dataset with unlabeled data,
            $B_s$: Batch size, 
            \hbox{$\theta_{g}$, $\theta_{d}$: Initial generator and discriminator parameters},
            $\mu$: Population size,
            $\lambda$: Offspring size,
            $\tau$: Tournament size,
            $n_e$: Number of evaluation batches,
            $n_t$: Number of training epochs per generator-discriminator couple
            \newline		
		\textbf{Return:}
		~$g,d$: Trained generator and discriminator
	}\label{alg:our-method}
	\begin{algorithmic}[1]
        \State $\mathbf{g} \gets$ initializePopulation($\theta_{g}$) \Comment{Initialize population $\mathbf{g}$}\label{alg:initialization-g}
        \State $\mathbf{d} \gets$ initializePopulation($\theta_{d}$) \Comment{Initialize population $\mathbf{d}$}\label{alg:initialization-d}
        \State $\iota \gets \left\lfloor\frac{T_B}{n_t  \lambda} \right\rfloor$ \Comment{Compute the number of generations to perform}
	\State $\mathcal{L}_{g,d} \gets$ evaluate($\mathbf{g}, \mathbf{d}, n_e$) \Comment{Evaluate populations} \label{alg:fit-eval-1}
	\For{$i=1$ \textbf{to} $\iota$} \Comment{Loop over generations}  
      \State $\mathbf{g}^* \gets$ select($\lambda, \tau, \mathcal{L}_{g,d}$) \Comment{Tournament selection to get offspring population $\mathbf{g}^*$} \label{alg:select-g}
      \State $\mathbf{d}^* \gets$ select($\lambda, \tau, \mathcal{L}_{g,d}$) \Comment{Tournament selection to get offspring population 
      $\mathbf{d}^*$} \label{alg:select-d} 
      \State $\mathbf{d}' \gets \mathbf{d}^*$ \Comment{Initialize the set of discriminators to couple with generators}
		\For{$g \in \mathbf{g}^*$} \Comment{Loop over $g$ in the offspring $\mathbf{g}^*$}
        \State $d \gets $ pickOneRandomly($\mathbf{d}'$) \Comment{Select a couple discriminator for $g$}
        \State $\mathbf{d}' \gets \mathbf{d}' - \{d\}$ \Comment{Remove the selected discriminator from the set}
		\State $g, d \gets$ train($g, d, n_t, p_{real,la}, p_{real,un}, B_s, \theta_g, \theta_d$) \Comment{Update $g$, $d$ parameters} \label{alg:method-ssl-training}
		\EndFor
            \State $\mathbf{g} \gets \mathbf{g} \cup \mathbf{g}^*$ \Comment{Add offspring to population} \label{alg:add-offspring-g}
            \State $\mathbf{d} \gets \mathbf{d} \cup \mathbf{d}^*$ \Comment{Add offspring to population} \label{alg:add-offspring-d}
		\State $\mathcal{L}_{g,d} \gets$ evaluate($\mathbf{g}, \mathbf{d}, n_e$) \Comment{Evaluate populations} \label{alg:fit-eval-2}
        \State $\mathbf{g} \gets$ updatePopulation($\mathbf{g}$) \Comment{Apply elitism (best $\mu$ individuals survive)} \label{alg:update-pops-g}
		\State $\mathbf{d} \gets$ updatePopulation($\mathbf{d}$) \Comment{Apply elitism (best $\mu$ individuals survive)} \label{alg:update-pops-d}
        \EndFor
        \State $g \gets$ selectBestGenerator($\mathbf{g}$)  \label{alg:select-best-g}
        \State $ d \gets$ selectBestDiscriminator($\mathbf{d}$)  \label{alg:select-best-d}
		\State \Return $g$, $d$ \Comment{Return $g$, $d$ trained SSL-GAN}\label{alg:return}
	\end{algorithmic}
\end{algorithm}

\section{Experimental setup}
\label{sec:experimental-setup}

We perform an empirical analysis of \method\ by examining both the discriminator classification accuracy and the quality of the samples produced by the generator\footnote{The replication package of the paper is at \url{https://doi.org/10.5281/zenodo.14729793}}. Our goal is to answer the following research questions:
\begin{itemize}
    \item \textbf{RQ1}: How do the number of training epochs during mutation, $n_t$, population size, $\mu$, and offspring size, $\lambda$, influence the performance of \method?
    \item \textbf{RQ2}: How does \method\ perform compared to the basic SSL-GAN?
    \item \textbf{RQ3}: How sensitive is the performance of \method\ and the basic SSL-GAN to the proportion of labeled data?
\end{itemize}

\begin{figure}
    \begin{subfigure}[t]{0.3\linewidth}
    \centering
    \includegraphics[width=\linewidth]{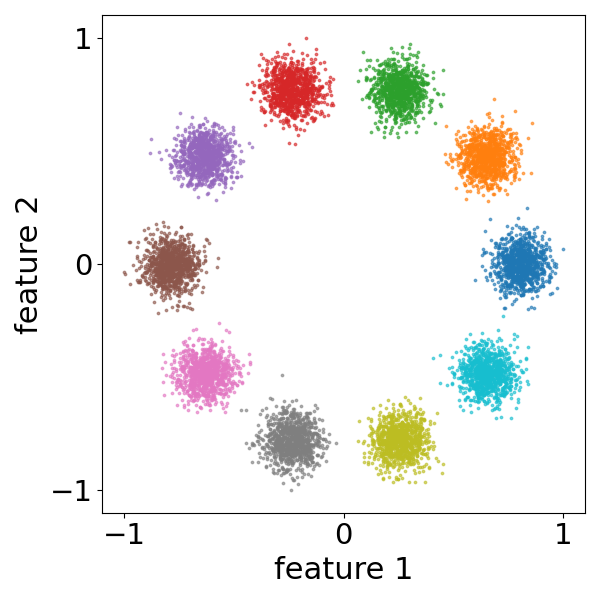}
    \caption{\RING dataset}
    \label{fig:ring-dataset}
\end{subfigure}
\begin{subfigure}[t]{0.3\linewidth}
    \centering
    \includegraphics[width=\linewidth]{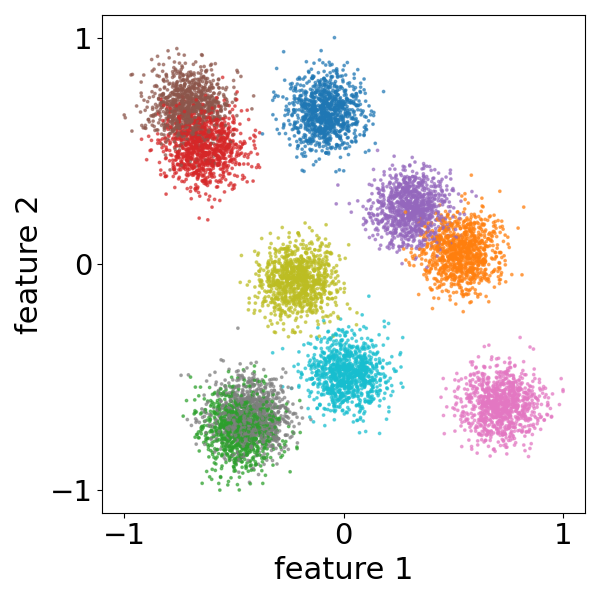}
    \caption{\BLOB dataset}
    \label{fig:blobs-dataset}
\end{subfigure}
\begin{subfigure}[t]{0.3\linewidth}
    \centering
    \includegraphics[width=\linewidth]{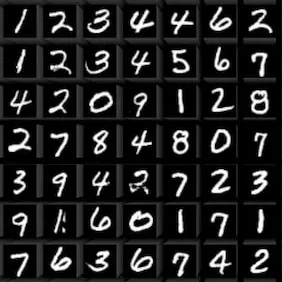}
    \caption{\MNIST dataset}
    \label{fig::mnist-dataset}
\end{subfigure}
    \caption{Representation of the datasets applied in our empirical analysis}
    \label{fig:datasets}
\end{figure}

The experiments are performed on three datasets: mixture of ten 2D Gaussian distributions arranged in a ring (referred to as \RING), a mixture of eight 2D Gaussian distributions randomly arranged (referred to as \BLOB), and the Modified National Institute of Standards and Technology (\MNIST) dataset~\cite{Deng2012141}.
Fig.~\ref{fig:datasets} illustrates the distributions of \RING and \BLOB and some samples of \MNIST.

\RING and \BLOB datasets consist of 2D points in the range [-1, 1] grouped in ten classes (Gaussian distributions), with 10,000 samples in the training set and 1,000 in the test set. These types of datasets are widely used in GANs research because, even if they may seem simple, they present a challenge in the generative component and can lead to mode collapse. 
The \RING dataset mixtures are well separated in the feature space, which makes it easier to classify the samples (see Fig.~\ref{fig:ring-dataset}).
In contrast, the mixtures in the \BLOB dataset are randomly distributed, with several mixtures sharing feature space, which means that a classifier cannot learn to classify all the points in the test set correctly. In our analysis, the BLOB dataset used in the whole experimentation is the same (see Fig.~\ref{fig:blobs-dataset}). A classifier was trained using all available labels in the training dataset, achieving a maximum classification accuracy of 0.889.

The \MNIST dataset contains gray-scale images of handwritten digits (0 to 9) with a resolution of 28$\times$28 pixels. The training set includes 60,000 images, while the test set contains 10,000 images.

The classification accuracy is calculated by applying the classifier defined by the discriminator on the test dataset. 
The generative models in \RING and \BLOB experiments are evaluated according to the Wasserstein distance ($W_D$)\cite{vallender1974calculation}, while in \MNIST are assessed regarding Fréchet Inception Distance (FID)~\cite{fid}.  

$W_D$ measures the minimal cost required to transform one probability distribution into another. Given two probability distributions $P$ and $Q$ on a metric space $\mathcal{X}$, $W_D(P, Q)$ is defined in Eq.~\eqref{eq:wd}, where $\Pi(P, Q)$ denotes the set of all joint distributions $\gamma(x, y)$ with marginals $P$ and $Q$, and $d(x, y)$ is the distance between points $x$ and $y$ in $\mathcal{X}$.

\begin{equation}
W_D(P, Q) = \inf_{\gamma \in \Pi(P, Q)} \int_{\mathcal{X} \times \mathcal{X}} d(x, y) \, d\gamma(x, y)
\label{eq:wd}    
\end{equation}

FID computes the similarity between two sets of images. Specifically, given two distributions of features extracted from ANNs, $P$ (e.g., real images) and $Q$ (e.g., generated images), FID is computed using Eq.~\eqref{eq:fid}, where $\mu_P$ and $\Sigma_P$ denote the mean and covariance of the features from $P$, and $\mu_Q$ and $\Sigma_Q$ denote those of $Q$.

\begin{equation}
\text{FID}(P, Q) = ||\mu_P - \mu_Q||^2 + \text{Tr}(\Sigma_P + \Sigma_Q - 2(\Sigma_P \Sigma_Q)^{1/2})
    \label{eq:fid}
\end{equation}

The experiments are performed on two different architectures for the SSL-GANs: one based on multilayer perceptrons (MLP) to learn \RING and \BLOB datasets, and one based on convolutional neural networks (CNN) to address \MNIST dataset. 

For \RING and \BLOB datasets, the generator consists of a three-layer MLP. It maps a latent vector to a 2D point (data sample) through a hidden layer with 64 units using ReLU activations. The final layer applies $\tanh$ activation to constrain the output range to $[-1, 1]$, fitting the \RING and \BLOB distribution. The discriminator is a two-layer MLP that maps a 2D input to a 64-dimensional feature space using LeakyReLU activations. It has two outputs: a binary classifier with a sigmoid activation to determine real or fake samples and a multiclass classification layer that outputs class predictions.

For \MNIST, the generator's input layer is fully connected and converts a latent vector into an initial 7$\times$7 feature map. This feature map is then upsampled through two layers of transposed convolutions, using ReLU activations and batch normalization, finally producing a 28$\times$28 output with a final activation function of tanh.
The discriminator architecture includes four convolutional blocks with LeakyReLU activations, dropout, and batch normalization, reducing the input image dimensions from 28$\times$28 (input image) to 4$\times$4 (features map). Then, it has two output layers: a binary classifier  (real/fake), which applies a sigmoid activation function, and a layer for multiclass classification of the MNIST digits. 

The experimental design focuses on assessing the impact of three key factors in a co-evolutionary algorithm (\textbf{RQ1}): population size, offspring size, and exploitation of individuals. We explore population sizes $\mu \in \{3, 5, 7, 9\}$, and adjust the offspring size $\lambda$ from 1 up to the ceiling of half the population size, e.g., $\lambda \in \{1, 2, 3\}$ for $\mu$=5. We examine the effect of the different levels of exploitation of solutions by varying the number of training epochs applied to update the network weights each generation ($n_t)$. We consider $n_t \in \{1, 5, 10\}$. 
This parameter is essential because the number of generations ($\iota$) that \method\ performs depends on the total training epochs to be conducted ($T_B$), $\lambda$, and $n_t$ (see~Eq.~\ref{eq:generations}). Thus, when $n_t$=1 \method\ completes 10 times more generations than when it is set $n_t$=10.
To evaluate the effectiveness of the proposed method, we compare it to a standard SSL-GAN training approach (\textbf{RQ2}), offering insights into the performance of the co-evolutionary algorithm across different parameter settings. 

The main hyper-parameters used are set according to preliminary experiments. Generators and discriminators apply the Adam optimizer with the same learning rate 0.0003. The batch size is 100 samples for \RING and \BLOB and 600 for \MNIST. The total training epochs $T_B$ is set to $300\lambda$.
We decided to set the budget proportional to the number of offspring to simulate the parallel training of the offspring. Thus, the standard SSL-GAN runs for 300 training epochs (i.e., $T$=$T_B$ because it is considered as $\lambda$=1).  


We denote with $n_s$ the number of labeled samples per class. For the \RING and \BLOB datasets, the experiments are performed with one labeled sample per class, $n_s=1$ (i.e., 10 labeled samples which is a portion of 0.001 labeled data in the training dataset). For the experiments on the \MNIST dataset, the primary analysis is performed using $n_s=10$ labeled samples per class (i.e., 100~labeled samples which is 0.00167 labeled data in the training dataset). A sensitivity analysis is performed on the number of labeled data for MNIST (\textbf{RQ3}), the most challenging dataset, to see the impact on the results. Thus, additional experiments are performed with $n_s=60$ and $n_s=100$ labeled samples per class.


\section{Experimental results}
\label{sec:results}

In this section, we present the results of the experiments, split by dataset. In each dataset, we first analyze the influence of the number of training epochs during offspring generation, $n_t$, and then the population and the size of the offspring, $\mu$ and $\lambda$, respectively (\textbf{RQ1}). We also show the results obtained by a standard SSL-GAN (\textbf{RQ2}). Thirty independent runs were performed in all the cases, except in the analysis of the sensitivity of the methods to the number of labels (\textbf{RQ3}) in \MNIST, where 15 independent runs were performed.

\subsection{\RING experimental results}
\label{subsec:ring-results}

\begin{figure}[!ht]
\centering
\begin{subfigure}[t]{0.45\linewidth}
    \centering
    \includegraphics[width=0.9\linewidth]{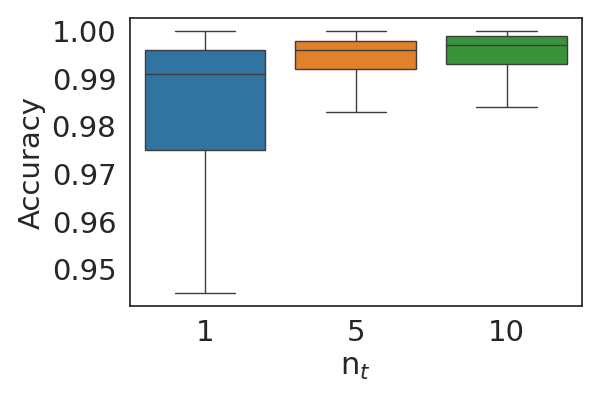}
    \caption{Classification accuracy}
    \label{fig:ring-dataset-accuracy}
\end{subfigure}%
\hfill
\begin{subfigure}[t]{0.45\linewidth}
    \centering
    \includegraphics[width=0.9\linewidth]{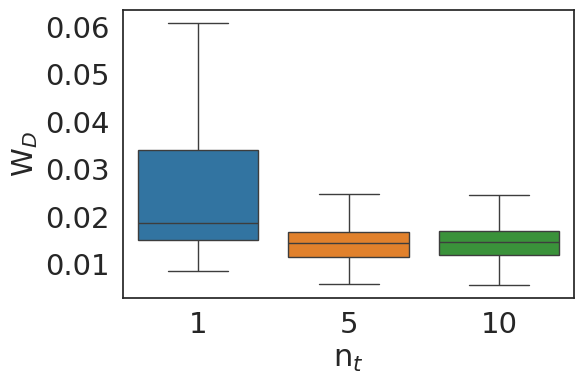}
    \caption{$W_D$}
    \label{fig:ring-dataset-wd}
\end{subfigure}

\begin{subfigure}[t]{0.45\linewidth}
    \centering
    \includegraphics[width=0.9\linewidth]{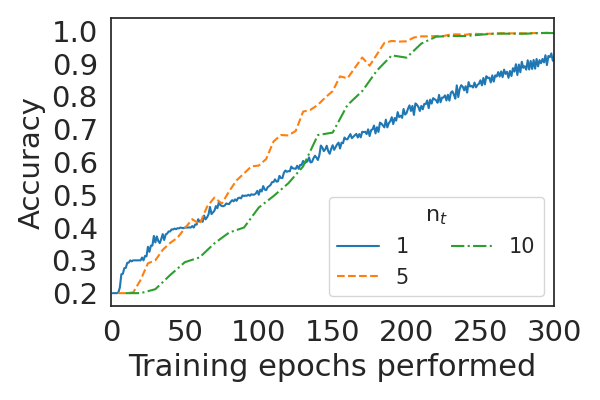}
    \caption{Median classification acc. evolution (first 300 training epochs)}
    \label{fig:ring-dataset-accuracy-evolution}
\end{subfigure}%
\hfill
\begin{subfigure}[t]{0.45\linewidth}
    \centering
    \includegraphics[width=0.9\linewidth]{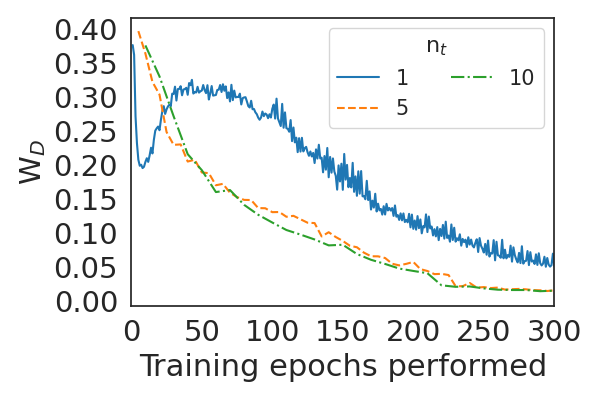}
    \caption{Median $W_D$ evolution (first 300 training epochs)}
    \label{fig:ring-dataset-wd-evolution}
\end{subfigure}
\caption{Influence of $n_t$ on \RING dataset}
\label{fig:ring-dataset-results}
\end{figure}

Table~\ref{tab:ring-dataset-results} shows the accuracy for the discriminator and the Wasserstein distance for the generator in the \RING dataset. We aggregate the results by value of $n_t$ to analyze the influence of that parameter. Figs.~\ref{fig:ring-dataset-accuracy}~and~\ref{fig:ring-dataset-wd} show the same results using boxplots. We can observe that the accuracy of the discriminator is higher when $n_t$ is 5 or 10, this means, more training epochs during the offspring generation. There is no statistically significant difference (at $\alpha=0.05$ level) between $5$ and $10$, but differences are statistically significant  between these two values and $n_t=1$. The same observations can be made for the generator ($W_D)$, where, again, $n_t=5$ and $n_t=10$ are statistically significantly better (lower value for $W_D$) than $n_t=1$. 

\begin{table}[!ht]
\centering
\caption{Classification accuracy of discriminator and Wasserstein distance of generator grouped by $n_t$ for \RING dataset.}
\label{tab:ring-dataset-results}
\begin{tabular}{r@{\hspace{1em}}r@{\hspace{1em}}r@{\hspace{1em}}r@{\hspace{1em}}rc@{\hspace{2em}}r@{\hspace{1em}}r@{\hspace{1em}}r@{\hspace{1em}}r}
\toprule
& \multicolumn{4}{c}{Accuracy} & & \multicolumn{4}{c}{$W_D$}\\
\cline{2-5} \cline{7-10}
$n_t$ & Min & Median & IQR & Max & & Min & Median & IQR & Max \\
\midrule
1 & 0.550 & 0.991 & 0.021 & 1.000  & & 0.009 & 0.019 & 0.019 & 0.668 \\
5 & 0.787 & 0.996 & 0.006 & 1.000  & & 0.006 & 0.014 & 0.005 & 0.110 \\
10 & 0.709 & 0.997 & 0.006 & 1.000 & & 0.006 & 0.015 & 0.005 & 0.111 \\
\bottomrule
\end{tabular}
\end{table}

Figs.~\ref{fig:ring-dataset-accuracy-evolution} and~\ref{fig:ring-dataset-wd-evolution} show the evolution of the median accuracy and the Wasserstein distance during the search in the experiments for different values of $n_t$. We can observe a faster convergence and a more stable evolution when $n_t$ is 5 and~10. The evolution for $n_t=1$ is noisy and slower.





\begin{figure}[!ht]
\begin{subfigure}[t]{0.9\linewidth}
    \centering
    \includegraphics[width=0.8\linewidth]{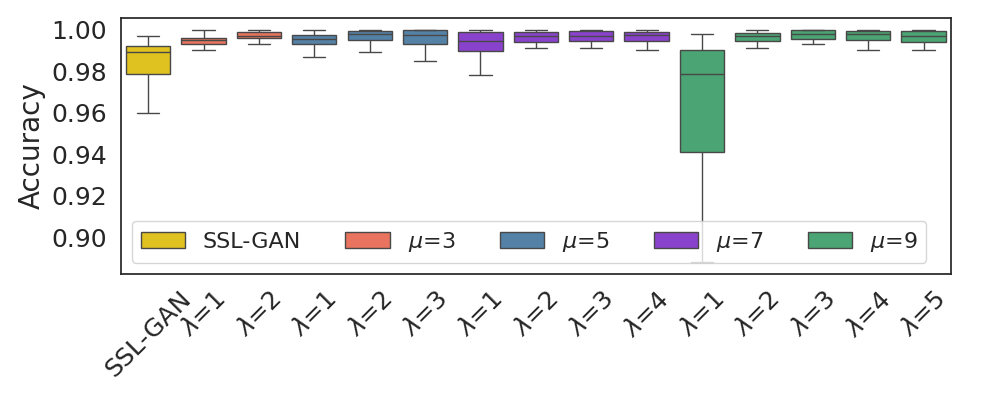}
    \caption{Classification results for \RING dataset when using $n_t$=10}
    \label{fig:ring-accuracy-population}
\end{subfigure}%

\begin{subfigure}[t]{0.9\linewidth}
   \centering
    \includegraphics[width=0.8\linewidth]{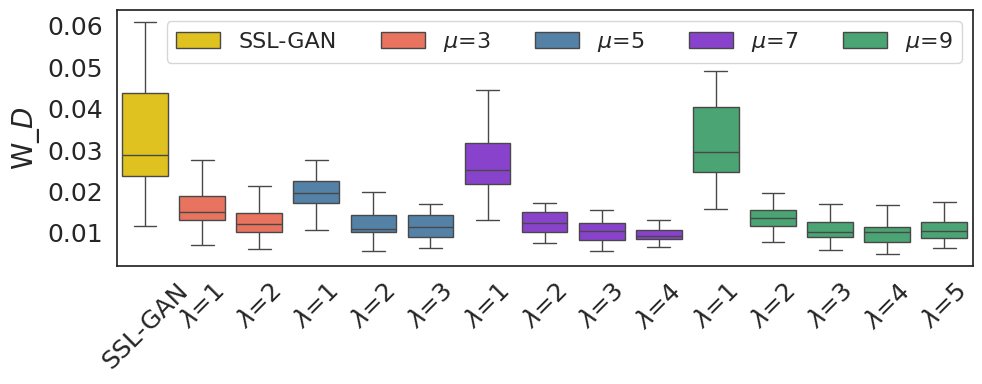}
    \caption{$W_D$ results for \RING experiments when using $n_t$=10}
    \label{fig:ring-wd-population}
\end{subfigure}
\caption{Influence of population and offspring size in the performance for \RING}
\label{fig:ring-populations}
\end{figure}

Now, we move to the influence of the population size, $\mu$, and offspring size, $\lambda$, on the performance of the networks. Figs.~\ref{fig:ring-accuracy-population} and~\ref{fig:ring-wd-population} show the accuracy of the discriminator and the Wasserstein distance of the generator for the \RING dataset for different combinations of $\mu$ and $\lambda$. Regarding the accuracy of the discriminators, we observe that the basic SSL-GAN has lower accuracy compared to \method. The differences are statistically significant except in the comparison with the \method variations with $\mu=7$, $\lambda=1$ and $\mu=9$, $\lambda=1$. There is no clear influence of the population size, $\mu$, on the accuracy. We can observe that the accuracy is higher when the number of offspring, $\lambda$, is greater than 1. The conclusion here is that we should set $\lambda > 1$.
Similar results, but with a more clear trend, can be observed in the case of the generator performance, using $W_D$. The generators of SSL-GAN have a higher value for $W_D$, compared to those of \method, with statistically significant differences. Here again, a value of $\lambda > 1$ decreases the Wasserstein distance. We can also appreciate a reduction in $W_D$ when $\mu$ is increased, but the statistical tests show no significant differences.
We conclude that \method\ outperforms the results of the basic SSL-GAN (\textbf{RQ2}), the population size, $\mu$, has no significant difference on the results, and the number of offsprings should be more than one (\textbf{RQ1}).

\begin{figure}
\begin{subfigure}[t]{0.49\linewidth}
    \centering
    \includegraphics[width=0.9\linewidth]{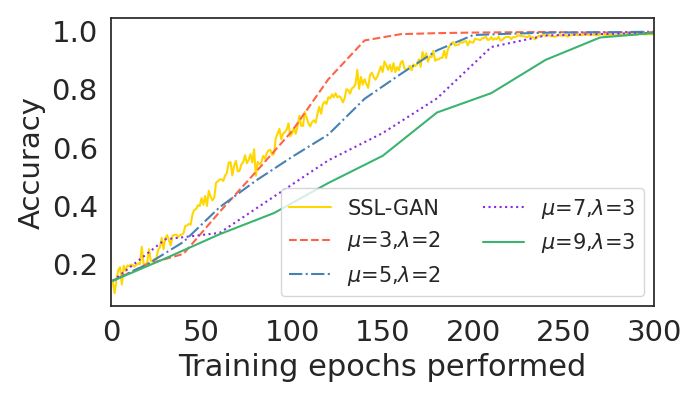}
    \caption{Median classification acc. evolution}
    \label{fig:ring-evolution-accuracy}
\end{subfigure}%
\hfill
\begin{subfigure}[t]{0.49\linewidth}
    \centering
    \includegraphics[width=0.9\linewidth]{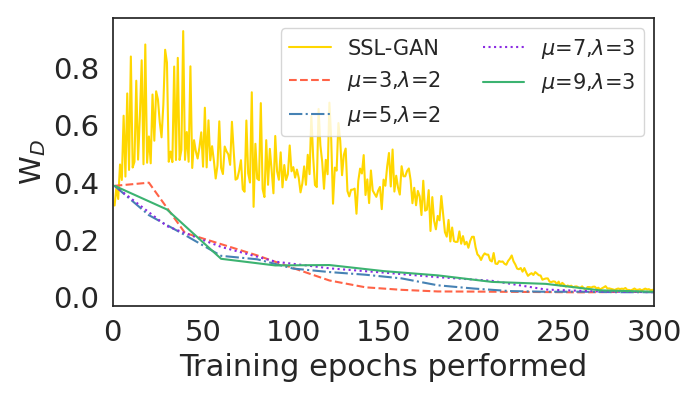}
    \caption{Median $W_D$ evolution}
    \label{fig:ring-evolution-wd}

\end{subfigure}
    \caption{Evolution of the classification accuracy and $W_D$ through each training epoch performed for the selected methods on \RING (first 300 training epochs)}
    \label{fig:gaussian_evolution}
\end{figure}

In Figs.~\ref{fig:ring-evolution-accuracy} and~\ref{fig:ring-evolution-wd} we show the evolution of the discriminator accuracy and the Wasserstein distance (for generators) in the \RING dataset for some selected configurations of \method\ and SSL-GAN. We can observe in both figures that the evolution of SSL-GAN is noisy and slow for $W_D$. \method, on the other hand, shows a soft evolution with a convergence time that depends on the population size, $\mu$, in the case of the accuracy. 
The fastest configuration of \method\ is the one with $\mu=3$. Increasing the population size also increases the time to reach the maximum accuracy.
Offspring's generation can be done in parallel. Parallelization would compress the curves to the left and they could overlap if we consider the evolution with respect to the wall clock time.




\subsection{\BLOB experimental results}
\label{subsec:blob-results}

\begin{figure}[!ht]
\centering

\begin{subfigure}[t]{0.45\linewidth}
    \centering
    \includegraphics[width=0.9\linewidth]{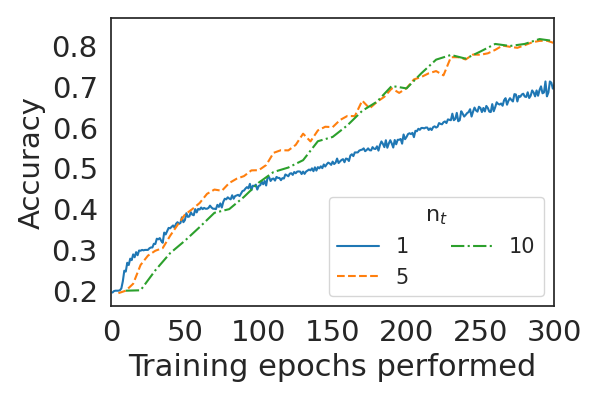}
    \caption{Median classification acc. evolution}
    \label{fig:blobs_accuracy_evolution_variation_training}
\end{subfigure}%
\hfill
\begin{subfigure}[t]{0.45\linewidth}
    \centering
    \includegraphics[width=0.9\linewidth]{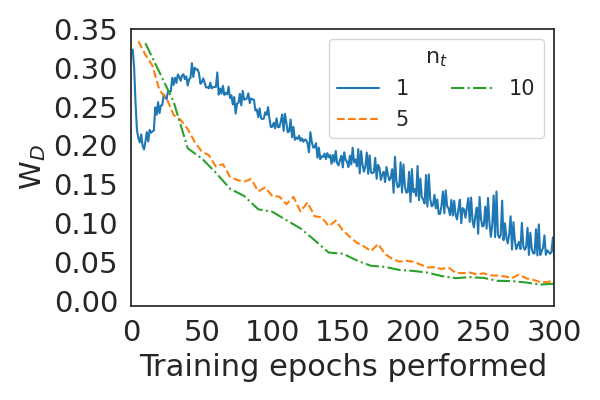}
    \caption{Median $W_D$ evolution}
    \label{fig:blobs_distance_evolution_variation_training}
\end{subfigure}
\caption{Influence of $n_t$ in the \BLOB dataset (first 300 training epochs)}
\label{fig:blobs_variation}
\end{figure}

The influence of $n_t$ in the discriminator accuracy and the Wasserstein distance obtained by the generators in the \BLOB dataset are presented in Table~\ref{tab:blob-dataset-results}. Figs.~\ref{fig:blobs_accuracy_evolution_variation_training} and~\ref{fig:blobs_distance_evolution_variation_training} shows the evolution of the median accuracy and median $W_D$ for the discriminator and generator, respectively, during the search. We observe the same behavior we saw in the \RING dataset: $n_t$ should be larger than 1, but there are no differences between 5 and 10 (\textbf{RQ1}).

\begin{table}[!ht]
\centering
\caption{Classification accuracy of discriminator and Wasserstein distance of generator grouped by $n_t$ for \BLOB dataset.}
\label{tab:blob-dataset-results}
\begin{tabular}{r@{\hspace{1em}}r@{\hspace{1em}}r@{\hspace{1em}}r@{\hspace{1em}}rc@{\hspace{2em}}r@{\hspace{1em}}r@{\hspace{1em}}r@{\hspace{1em}}r}
\toprule
& \multicolumn{4}{c}{Accuracy} & & \multicolumn{4}{c}{$W_D$}\\
\cline{2-5} \cline{7-10}
$n_t$ & Min & Median & IQR & Max & & Min & Median & IQR & Max \\
\midrule
1  & 0.473 & 0.826 & 0.040 & 0.861 && 0.006 & 0.018 & 0.021 & 0.449 \\
5  & 0.516 & 0.829 & 0.012 & 0.861 && 0.006 & 0.013 & 0.009 & 0.291 \\
10 & 0.568 & 0.830 & 0.011 & 0.857 && 0.005 & 0.013 & 0.008 & 0.075 \\
\bottomrule
\end{tabular}
\end{table}


\begin{figure}[!ht]
\centering
\begin{subfigure}[t]{0.9\linewidth}
\centering
    \includegraphics[width=0.7\linewidth]{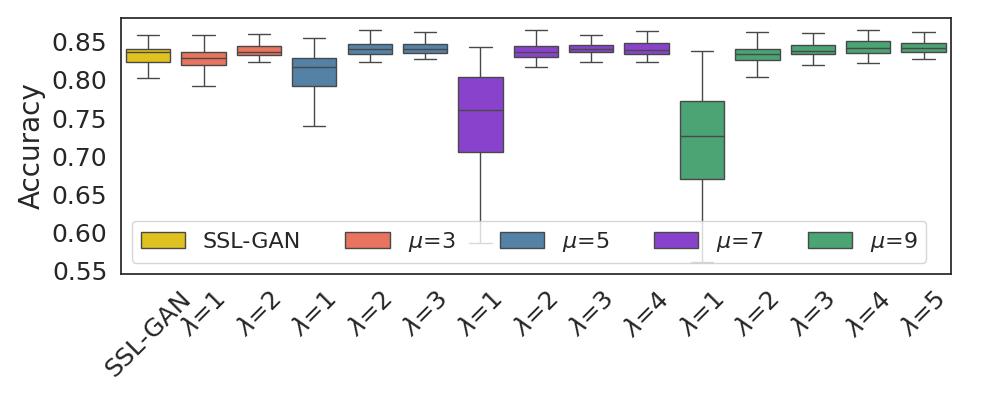}
    \caption{Classification results for \BLOB experiments when using $n_t$=10}
    \label{fig:blob-accuracy-population}
\end{subfigure}

\begin{subfigure}[t]{0.9\linewidth}
\centering
    \includegraphics[width=0.7\linewidth]{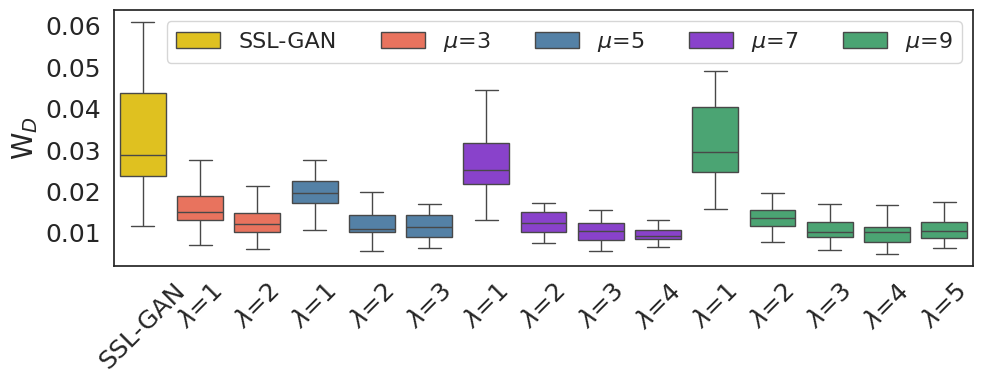}
    \caption{$W_D$ results for \BLOB experiments when using $n_t$=10}
    \label{fig:blob-wd-population}
\end{subfigure}    
\caption{Influence of population and offspring size in the performance for \BLOB}
\label{fig:blob-figures}
\end{figure}



Regarding the influence of the population size and the number of offspring per generation, we also observe the same trends as in the \RING dataset in Figs.~\ref{fig:blob-accuracy-population} and~\ref{fig:blob-wd-population}: \method\ outperform the results of SSL-GAN (\textbf{RQ2}), and the best performance is obtained when $\lambda > 1$, while $\mu$ does not seem to have much influence on the results (\textbf{RQ1}).

\begin{figure}
\begin{subfigure}[t]{0.49\linewidth}
    \centering
    \includegraphics[width=0.9\linewidth]{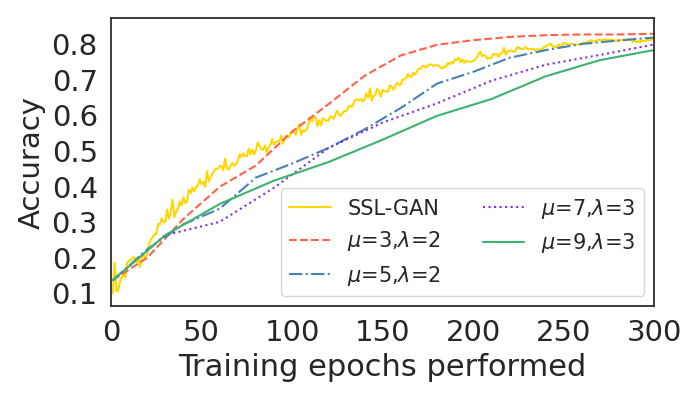}
    \caption{Median classification acc. evolution}
    \label{fig:evolution-accuracy}
\end{subfigure}%
\hfill
\begin{subfigure}[t]{0.49\linewidth}
    \centering
    \includegraphics[width=0.9\linewidth]{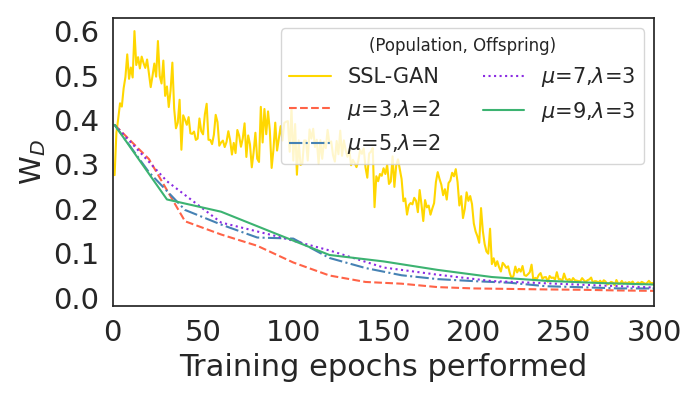}
    \caption{Median $W_D$ evolution}
    \label{fig:evolution-fid}

\end{subfigure}
    \caption{Evolution of the classification accuracy and $W_D$ through each training epoch performed for the selected methods (first 300 training epochs) for \BLOB}
    \label{fig:blobs_evolution}
\end{figure}






\subsection{\MNIST experimental results}
\label{sec:mnist-results}

\begin{figure}
\setlength{\abovecaptionskip}{5pt} 
\setlength{\belowcaptionskip}{0pt} 

\begin{subfigure}[t]{0.49\linewidth}
\setlength{\abovecaptionskip}{5pt} 
\setlength{\belowcaptionskip}{0pt} 
    \centering
    \includegraphics[width=0.9\linewidth]{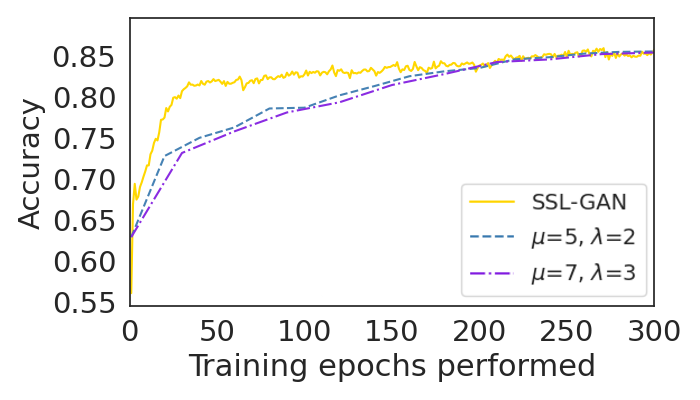}
    \caption{Median classification acc. evolution}
    \label{fig:mnist-evolution-accuracy}
\end{subfigure}
\hfill
\begin{subfigure}[t]{0.49\linewidth}
\setlength{\abovecaptionskip}{5pt} 
\setlength{\belowcaptionskip}{0pt} 
    \centering
    \includegraphics[width=0.9\linewidth]{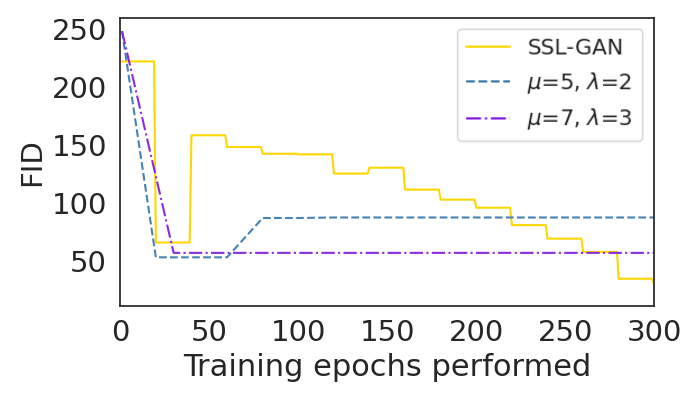}
    \caption{Median FID evolution}
    \label{fig:mnist-fid-evolution}
\end{subfigure}
    \caption{Evolution of the classification accuracy and FID through each training epoch performed for the selected methods (first 300 training epochs) for \MNIST}
    \label{fig:mnist-final}
\end{figure}

Taking into account the trends observed in the results for \RING and \BLOB, we only show results of \method\ for $n_t=10$ and two combinations of population and offspring size: $\mu$=5, $\lambda$=2; and $\mu$=7, $\lambda$=3. In Table~\ref{tab:mnist-results}
we show the statistics of the classification accuracy of the discriminators and the FID obtained by the generators. The main conclusion from these results is that \method\ outperforms the results of SSL-GAN (\textbf{RQ2}). The differences observed in the results between SSL-GAN and \method\ are statistically significant. The differences between the two population configurations ($\mu=5, \lambda=2$ and $\mu=7, \lambda=3$) are not statistically significant. In Figs.~\ref{fig:mnist-evolution-accuracy} and~\ref{fig:mnist-fid-evolution} we present the evolution of the accuracy and FID during the search. We can observe that the SSL-GAN is very fast at the beginning, it reaches better values for the accuracy and FID than \method\ in the first 300 training epochs. However, in the last 300 training epochs (not shown) \method\  outperforms SSL-GAN.



\begin{table}[!ht]
\centering
\caption{Classification accuracy of discriminator and FID of generator grouped for \method\ variations and SSL-GAN addressing \MNIST dataset.}
\label{tab:mnist-results}
\begin{tabular}{r@{\hspace{1em}}r@{\hspace{0.5em}}r@{\hspace{0.5em}}r@{\hspace{0.5em}}rc@{\hspace{1.5em}}r@{\hspace{0.5em}}r@{\hspace{0.5em}}r@{\hspace{0.5em}}r}
\toprule
&\multicolumn{4}{c}{Accuracy} & & \multicolumn{4}{c}{FID}\\
\cline{2-5} \cline{7-10}
& Min & Median & IQR & Max & & Min & Median & IQR & Max \\
\midrule
\method\ \\
{\scriptsize $\mu=5,\lambda=2$} & 0.759 & 0.883 & 0.012 & 0.898 && 22.590 & 27.134 & 14.744 & 44.734 \\
{\scriptsize $\mu=7,\lambda=3$} & 0.759 & 0.880 & 0.013 & 0.897 && 22.570 & 24.952 & 16.663 & 60.253 \\
\midrule
SSL-GAN & 0.751 & 0.857 & 0.020 & 0.879 && 24.769 & 29.335 & 34.140 & 96.689 \\
\bottomrule
\end{tabular}
\end{table}

We analyze in the \MNIST dataset what is the sensitivity of SSL-GAN and \method\ to the number of labels in the dataset (\textbf{RQ3}). We evaluate the accuracy of the discriminator when the number of samples per class $n_s$ in the labeled data is 10 (the base case), 60, and 100. Table~\ref{tab:mnist-accuracy-sensitivity} shows the median accuracy and interquartile range (IQR). We observe that as $n_s$ increases the accuracy also increases for all the methods. \method\ is still better than SSL-GAN in all the cases. However, the difference in the accuracy is decreasing between the two  methods with a larger number of samples per class.




\section{Conclusions and future work}
\label{sec:conclusions}

In this paper, we have proposed an elitist co-evolutionary algorithm for SSL-GAN training with a panmictic population. Our goal was to analyze the influence of different hyper-parameters of the method in the performance of the generators and discriminators. We can answer \textbf{RQ1} by concluding  that having more than one individual in the population ($\mu > 1$) and generating more than one individual in each generation ($\lambda > 1$) is beneficial for the search. We did not observe a clear influence of the population size $\mu$ on the results when $\mu > 1$. We also observe that our \method\ method outperforms a standard SSL-GAN, answering \textbf{RQ2}. Finally, we answer \textbf{RQ3} concluding that increasing the number of labels per class, $n_s$, improves the performance of both \method\ and SSL-GAN, and the differences between them are reduced when $n_s$ increases. Although our conclusions are only for the three datasets used and more experiments should be done to confirm these findings, we think the results in this paper could be used as a guide for parameter tuning of SSL-GAN training using EAs.

\begin{table}[!ht]
    \centering
\caption{Classification accuracy of discriminator for \method\ and SSL-GAN in the \MNIST dataset when the number of samples per class $n_s$ changes.}
    \label{tab:mnist-accuracy-sensitivity}
\renewcommand{\arraystretch}{0.9} 
\begin{tabular}{rr@{\hspace{1em}}r@{\hspace{1em}}rc@{\hspace{2em}}r@{\hspace{1em}}rc@{\hspace{2em}}r@{\hspace{1em}}r}
\toprule
 & & \multicolumn{2}{c}{$n_s=10$} && \multicolumn{2}{c}{$n_s=60$} && \multicolumn{2}{c}{$n_s=100$}\\
\cline{3-4} \cline{6-7} \cline{9-10} 
&  & \phantom{0}Median & IQR && \phantom{0}Median & IQR && \phantom{0}Median & IQR\\
\midrule
\method \\
\scriptsize{$\mu=5,\lambda=2$}   & & 0.883 & 0.012 && 0.953 & 0.002 && 0.956 & 0.011 \\
\scriptsize{$\mu=7,\lambda=3$}    & & 0.880 & 0.013 && 0.952 & 0.003  && 0.956 & 0.010 \\
\midrule
SSL-GAN & & 0.857 & 0.020 && 0.939 & 0.008 && 0.950 & 0.009 \\
\bottomrule
\end{tabular}
\vspace{-0.5cm}
\end{table}

Although we treat the training problem as a single-objective problem, discriminator training could be formulated using a multi-objective approach, where supervised and unsupervised losses, are considered as two different objectives of the training process. Besides, \method\  could be implemented using parallelism.
In addition, it would be interesting to analyze the influence of the spatial distribution of the population in the performance of the GAN. This could be done comparing the results of \method\ with that of Lipizzaner~\cite{DBLP:journals/asc/ToutouhNHO23}.

\subsection*{Acknowledgements}
This work is partially funded by the Junta de Andalucia, Spain, under contract QUAL21 010UMA; and Universidad de Málaga under the grant B1-2022\_18. We thankfully acknowledge the computer resources,  technical expertise and assistance provided by the SCBI center of the University of Malaga.

\bibliographystyle{splncs04}

\end{document}